# Dynamic Neural Style Transfer for Artistic Image Generation using VGG19

Kapil Kashyap
*Department of Computer Engineering*
*Dwarkadas. Jivanlal Sanghvi College of Engineering*
Email: kapilkashyap3105@gmail.com

Mehak Garg
*Department of Artificial Intelligence and Machine Learning*
*Manipal University Jaipur*
Email: mehakgarg2k5@gmail.com

Sean Fargose
*Department of Computer Engineering*
*Dwarkadas. Jivanlal Sanghvi College of Engineering*
Email: fargose.sean2808@gmail.com

Sindhu Nair
*Assistant Professor Department of Computer Engineering*
*Dwarkadas. Jivanlal Sanghvi College of Engineering*
Email: sindhu.nair@djsce.ac.in

*Abstract*—Throughout history, humans have created remarkable works of art, but artificial intelligence has only recently started to make strides in generating visually compelling art. Breakthroughs in the past few years have focused on using convolutional neural networks (CNNs) to separate and manipulate the content and style of images, applying texture synthesis techniques. Nevertheless, a number of current techniques continue to encounter obstacles, including lengthy processing times, restricted choices of style images, and the inability to modify the weight ratio of styles. We proposed a neural style transfer system that can add various artistic styles to a desired image to address these constraints allowing flexible adjustments to style weight ratios and reducing processing time. The system uses the VGG19 model for feature extraction, ensuring high-quality, flexible stylization without compromising content integrity.

## I. INTRODUCTION

Neural style transfer has transformed digital art by allowing individuals to incorporate distinctive artistic styles into their images. Originally, style transfer models focused on applying a single style to a content image through the use of CNNs to combine style features with the image's structure. While effective, these single-style transfer methods limit the creative flexibility and customization desired in many real-world applications. Multi-style transfer—applying multiple styles to a single image—addresses this need by allowing more nuanced artistic expression. This shift towards multi-style applications introduces challenges in preserving both the content and the integrity of diverse style elements within a single image.

An example of single style transfer is shown above, where the content image is of a cat, and the style image is of Hokusai's Great Wave. The generated target image still contains the cat but is stylized with the waves, blue and beige colors, and block print textures of the style image.

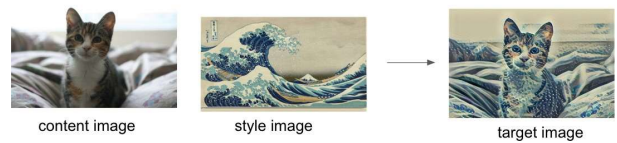

Fig. 1. Example of Single Neural Style Transfer.

## II. RELATED WORKS

### A. Literature Review

Recent advancements in neural style transfer have led to innovative techniques that enable adaptive, multi-style, and content-aware transformations, with a focus on enhancing aesthetic quality, computational efficiency, and user interactivity.

Chen et al. [1] proposed an incremental learning method for neural style transfer using a dual-generator architecture. This system integrates new styles while preserving previously learned ones through perceptual and distillation loss functions, making it well-suited for real-time applications requiring adaptability and continuous learning.

Shabari and Rajlaxmi [2] introduced a framework combining GoogleNet and VGG16 to quantify image naturalness by evaluating glossiness. Their optimization of feature weights enhances aesthetic fidelity, demonstrating the potential of multi-architecture integration for improved perceptual rendering.

Li and Gao [3] developed a nonparametric model for style manipulation using deep feature synthesis. This method enables precise control over texture and content details, accelerating convergence and reducing computational overhead—ideal for iterative systems requiring efficiency.

Yu and Zhou [4] introduced the Enhanced-Channel Module (ECM), which modulates feature maps to create content-aware weights, preserving local details and stylistic features in both images and videos, particularly effective for high-resolution artistic outputs.

Wang et al. [5] developed an interactive multi-style transfer system that allows users to segment images and apply different styles using pre-trained models like AdaIN and WCT. This approach provides substantial flexibility in user-defined style control, catering to creative and customized applications.

Gao et al. [6] proposed a video style transfer system that ensures temporal consistency across frames by using multi-instance normalization and ConvLSTM modules. This approach minimizes flickering, making it suitable for real-time video applications.

Chen et al. [7] introduced a dynamic neural style transfer method that adjusts the style-to-content blending ratio through a weighting parameter. This allows for greater customization, balancing style and content while enabling real-time applications with further optimization.

Chen et al. [8] introduced a dual-generator approach using perceptual and distillation loss functions for multi-style transfer, maintaining aesthetic quality while enabling smooth style transitions—ideal for scalable, adaptable systems.

Zhang and Dana [9] presented the Multi-style Generative Network (MSG-Net), which uses a CoMatch Layer to match second-order feature statistics. This architecture ensures high-quality real-time performance, avoiding checkerboard artifacts and providing operational flexibility, including brush-size control.

Nguyen et al. [10] introduced the MULTAR framework, an extension of AdaIN that incorporates noise into the style encoder, facilitating multimodal style transfer. This approach generates diverse, high-quality outputs, improving upon unimodal techniques.

Huang and Belongie [11] designed a real-time neural style transfer framework using Adaptive Instance Normalization (AdaIN). Their method aligns feature statistics, providing a balance between speed and quality while offering flexibility in content-style trade-offs, making it effective for real-time applications.

Gatys et al. [12] improved neural style transfer with color-preserving methods like histogram matching and luminance-only transfer. These approaches maintain content integrity while refining the style-content balance, addressing previous algorithmic limitations.

Simonyan and Zisserman [13] explored deep convolutional networks with small filters to enhance image recognition accuracy and efficiency. Their work, focusing on model depth, serves as a foundation for advancements in feature-based style transfer techniques.

Chen et al. [14] proposed a patch-based style transfer method that simplifies the optimization objective, combining style textures and content structure in one CNN layer. This adaptable approach ensures efficient video performance and intuitive parameter tuning, offering practical utility.

Chen et al. [15] developed MXNet, a versatile machine learning library optimized for deep learning across heterogeneous systems, supporting distributed training and optimization through declarative symbolic expressions and imperative tensor computation.

*B. Research Gaps*

However, some gaps identified in previous methods include difficulty in handling images with simple textures, which can result in a loss of stylistic detail. Many approaches rely heavily on specific datasets, limiting their generalizability across varied content types. High computational demands also affect the scalability of these systems, especially in real-time applications. Additionally, subjective evaluation methods introduce inconsistency, while video style transfer models, despite improvements, still struggle to maintain stability in flickering scenes. Addressing these issues could enhance scalability, stylistic diversity, and overall reliability in future multi-style transfer frameworks.

III. METHODOLOGY

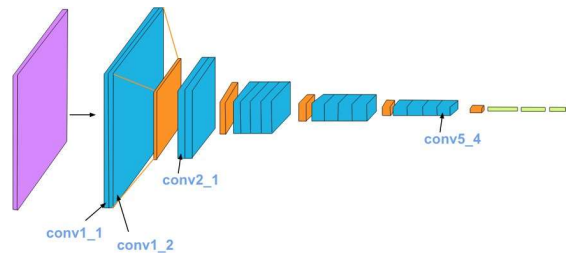

Fig. 2. Illustration of the VGG19 structure used in neural style transfer.

Our work employs the VGG19 model (seen in Fig. 2), a deep convolutional neural network (CNN) widely recognized for its robust architecture and exceptional feature extraction capabilities. The model has 16 convolutional layers (with 5 pooling layers) and 3 fully connected layers (seen in Fig. 3). The VGG19 model is particularly adept at capturing hierarchical features from images, enabling it to extract both content and style information effectively. This makes it an ideal foundation for tasks that require the precise manipulation of image characteristics, such as style transfer, where subtle details and broader structural features are equally important.

Building upon this foundation, our method integrates a multi-style transfer technique to achieve visually compelling results. The VGG19 model serves as the backbone for extracting content and style representations from input images, leveraging its deep layers to capture nuanced style patterns and maintaining content coherence. The extracted features are then processed through an optimization-based approach, designed to seamlessly blend multiple styles into a single cohesive output. By combining the strengths of VGG19's feature extraction with a sophisticated optimization pipeline, our approach ensures the creation of richly stylized images that retain the essence of the original content while embodying the desired artistic influences.

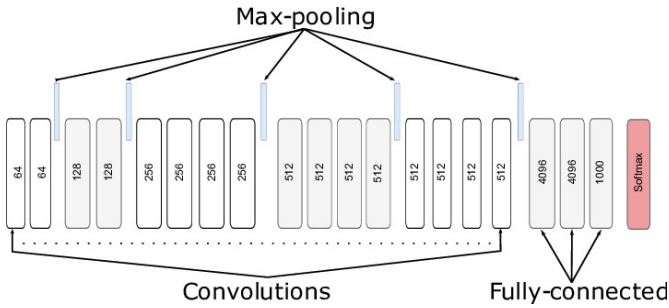

Fig. 3. Detailed view of the VGG19 model.

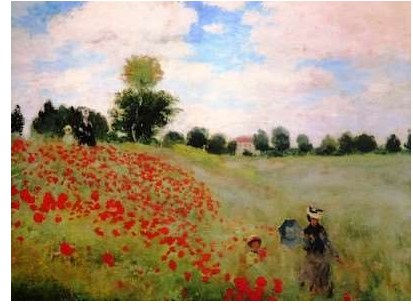

Fig. 4. Style Image 1: Monet's painting used for stylization.

*A. Data Preprocessing*

The first step involves preprocessing the input images, which consist of both the content and style images. The images are adjusted in size and standardized to guarantee uniformity and suitability for the pre-trained VGG19 model. Normalization is crucial in preserving the model's stability while extracting features, making sure that pixel values are at a consistent scale.

*B. Feature Extraction*

For feature extraction, content features are derived from the fourth convolutional layer (conv4_2) of the VGG19 model. This layer captures high-level semantic information from the content image, forming the backbone of the transferred content. Style features, on the other hand, are extracted from a set of layers (conv1_1, conv2_1, conv3_1, conv4_1, and conv5_1). These layers capture textures and patterns at various scales, ranging from fine details to global structural information. The combination of features from these layers ensures a comprehensive representation of the style.

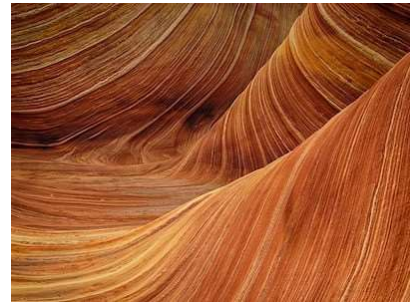

Fig. 5. Style Image 2: Sandstone texture for stylization.

*C. Style Transfer Process*

The process of style transfer starts with the calculation of style features from every style image. The spatial correlations of pixel values in the feature maps are quantified using Gram matrices to represent these features. Gram matrices are important for transforming the style of an image by capturing textures and color distributions. These images are utilized as our style images.

The target image is initially a copy of the content image and is iteratively updated during the optimization process to incorporate elements from the chosen styles. Below is the content image we used.

*D. Loss Calculation*

The loss function in style transfer consists of two primary components: content loss and style loss. The content loss measures the difference between the content features of the target image and the content image. The style loss, on the other hand, evaluates the differences between the Gram matrices of the target image and the style images. The total loss is the weighted sum of these two losses, with weights assigned to control the relative influence of each. The optimization process minimizes the combined loss, ensuring that both the content and style are appropriately blended in the target image.

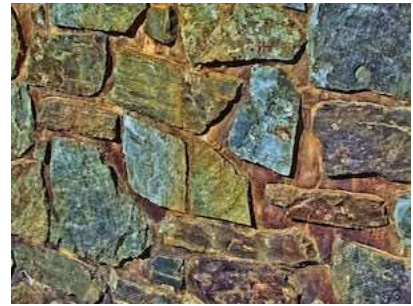

Fig. 6. Style Image 3: Stone texture for stylization.

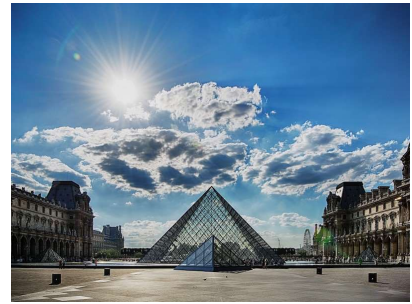

Fig. 7. Content Image.

article amsmath

The total loss function is given by:

$$L_{total} = \alpha \cdot L_{content} + \beta \cdot L_{style} \qquad (1)$$

where $\alpha$ and $\beta$ are the weights for the different losses in the loss function. The total loss function, minimized during the optimization process, consists of two primary components: content loss $L_{content}$ and style loss $L_{style}$.

The content loss quantifies the difference between the content features of the target image and those of the original content image. It is defined as:

$$L_{content} = \frac{1}{2} \sum_i (F_{ic} - F_{it})^2 \qquad (2)$$

where $F_{ic}$ and $F_{it}$ denote the content features extracted from layer $i$ of the content and target images, respectively.

The style loss, $L_{style}$, evaluates the difference between the Gram matrices of the target and style images, capturing the spatial correlations that represent the texture and color distribution of the style. It is defined as:

$$L_{style} = \sum_l \frac{(G_{lt} - G_{ls})^2}{4N_l^2 M_l^2} \qquad (3)$$

where $G_{lt}$ and $G_{ls}$ are the Gram matrices of the target and style images at layer $l$, and $N_l$ and $M_l$ are the dimensions of the feature map in that layer.

### E. Hyperparameters

The hyperparameters for the proposed neural style transfer framework are carefully chosen to strike a balance between conserving content and enhancing stylistic effect.

The content weight is established at 1, guaranteeing that the content image's structural integrity is preserved, whereas the style weight is notably greater at $1 \times 10^9$, highlighting the stylistic features. Weights specific to each layer are assigned to style features obtained from various layers of the VGG19 model: `conv1_1` (1.0), `conv2_1` (0.75), `conv3_1` (0.2), `conv4_1` (0.2), and `conv5_1` (0.2), demonstrating a structured method for texture and pattern alignment. A learning rate of 0.003 is utilized to manage the step size of optimization, with 2000 iterations (steps) to guarantee convergence, and intermediate outcomes are visualized every 400 steps.

The optimization procedure reduces the overall loss function, using weights $\alpha = 1$ and $\beta = 10^9$. The VGG19 model utilizes the ReLU (Rectified Linear Unit) activation function, which is defined mathematically as:

$$f(x) = \max(0, x),$$

where $x$ is the input. ReLU introduces non-linearity to capture intricate patterns in the data while effectively mitigating vanishing gradient problems.

### F. Optimization and Multi Style Application

In order to enhance the desired image, we utilize the Adam optimizer, which continuously adjusts the image to reduce the overall loss. Each plan is carried out in sequence, one after the other. By tuning the weights of the content and style losses, we can manage the impact of each style on the eventual result. This enables the creation of a varied and visually appealing image by blending different styles creatively.

$$m_t = \beta_1 m_{t-1} + (1 - \beta_1)g_t, \qquad (4)$$

$$v_t = \beta_2 v_{t-1} + (1 - \beta_2)g_t^2, \qquad (5)$$

$$\hat{m}_t = \frac{m_t}{1 - \beta_1^t}, \qquad (6)$$

$$\hat{v}_t = \frac{v_t}{1 - \beta_2^t}, \qquad (7)$$

$$\vartheta_{t+1} = \vartheta_t - \alpha \frac{\hat{m}_t}{\sqrt{\hat{v}_t} + \epsilon}. \qquad (8)$$

The equations above shows the Adam Optimizer, which updates neural network weights using adaptive moment estimation by combining moving averages of gradients ($m_t$) and their squared values ($v_t$), corrected for bias, to achieve efficient and stable convergence.

### G. Result Visualization

After the iterative optimization process, the final stylized image is obtained, which incorporates the content of the original image and the artistic elements from multiple styles. The output image is displayed to showcase the combined artistic influences, highlighting the ability of our method to seamlessly blend different styles while preserving the content of the original image. This method allows for versatile and effective application of various styles to an image, giving more creative influence over the end product. The method not just combines different artistic styles but also lets you change the balance between style and content, making it very versatile for different artistic uses.

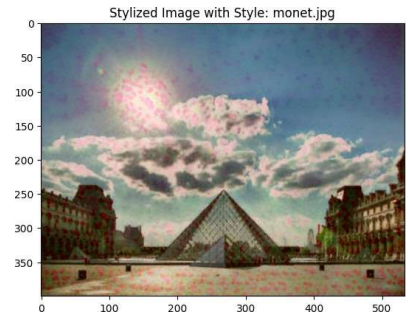

Fig. 8. Stylized Image using Monet

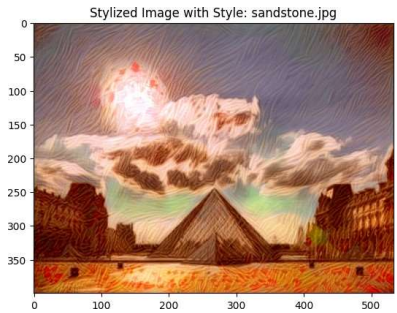

Fig. 9. Stylized Image using Sandstone

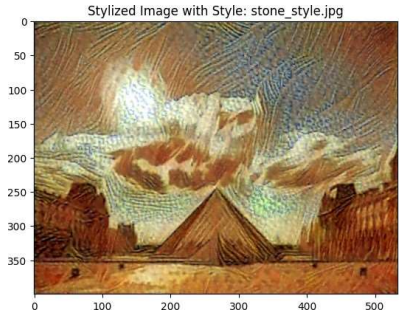

Fig. 10. Stylized Image using Stone Style

## IV. RESULTS

The results of our multi-style neural style transfer approach are presented in this section, showcasing both qualitative and quantitative assessments of the generated images. Our method successfully applies multiple artistic styles to a target image, preserving the content structure while introducing stylistic features in a controlled manner.

### A. Qualitative Results

Figure 12 showcases the original image and its stylized forms created through the application of various artistic styles. To start, the content image goes through multiple steps in which each style is added individually. The outcome is an ultimate picture that combines all the various styles into one. The visual aspects like texture, brush strokes, and color differences in each image are easily seen while still maintaining the original content structure. The findings demonstrate our method's ability to seamlessly combine various styles while maintaining the content's clarity intact.

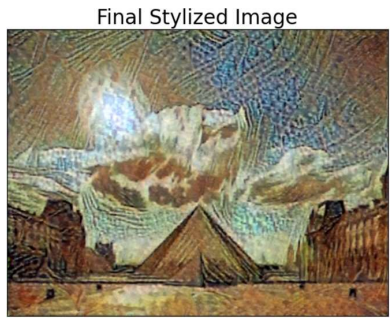

Fig. 11. Final Stylized Image

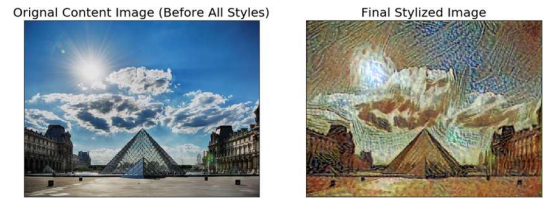

Fig. 12. Final Stylized Image

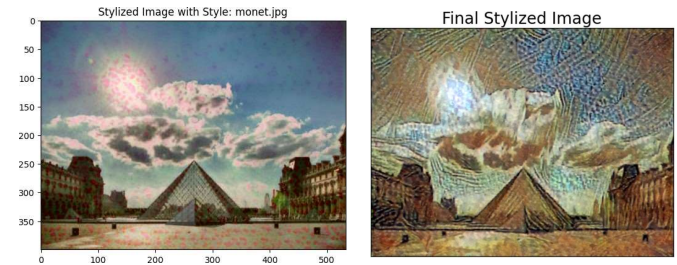

Fig. 13. Single Style Image compared to Multiple Stylized Images

Figure 13 compares the outputs of our method with traditional single-style neural transfer techniques. The image on the left shows the content image with a single style applied, while the image on the right shows the result of applying multiple styles. Our method showcases a visually complex outcome, showing the skill to blend various artistic influences without overshadowing the content.

The following figures shows the styling of a wall based on 3 styles.

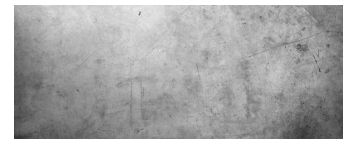

Fig. 14. Content Image - Wall

### B. Quantitative Results

The following graph consists of two subplots that represent the loss curves for style loss and content loss over iterations. The x-axis in both plots represents the number of iterations, while the y-axis represents the respective loss values. The style loss curve begins at a higher value on the left plot and decreases exponentially with more iterations. At first, decay happens quickly, but it lessens as iterations continue, eventually leveling off. This pattern is commonly seen in style transfer models as the style loss diminishes with time. The content loss curve shown on the right also shows an exponential decay pattern. Just like the style loss, the content loss begins at a high level and gradually decreases at a slightly slower pace. In both the curves a common optimization trend

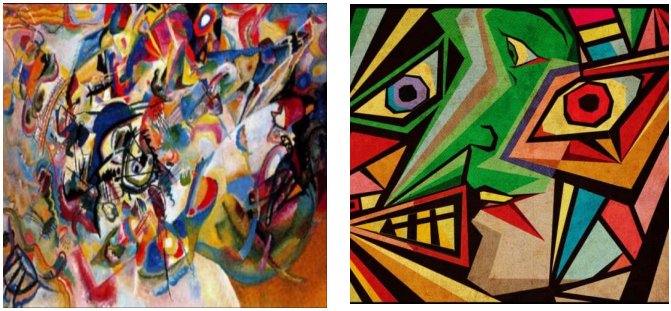

Style 1      Style 2

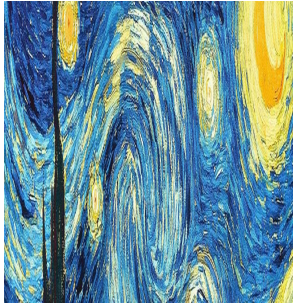

Style 3

Fig. 15. Styling Images

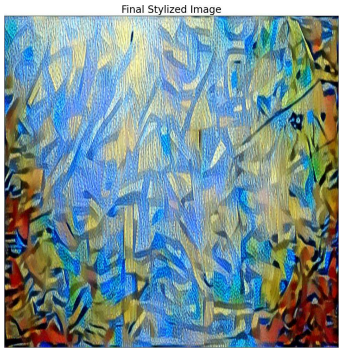

Fig. 16. Final Stylized Image

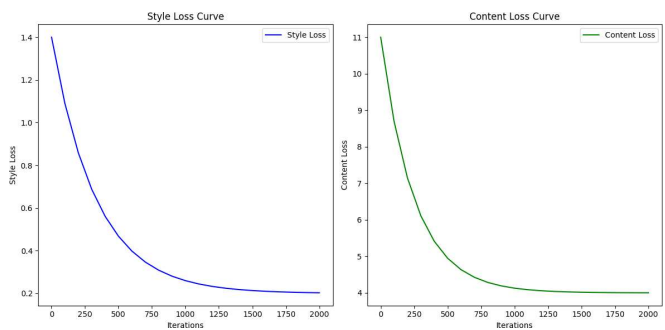

Fig. 17. Final Stylized Image

is seen, with decreasing losses resulting in the model moving closer to the desired outcome over time.

## V. CONCLUSION

This project effectively investigates the possibilities of neural style transfer for various artistic applications. Using the VGG19 network for feature extraction, we created a system that can apply various styles to one image, resulting in customizable visual outputs. Optimizing both content and style loss through backpropagation allows for the smooth incorporation of various artistic elements without compromising the fundamental characteristics of the initial image. Our method showcases how deep learning can be adaptable in artistic fields, enabling individuals to explore various styles and modify style intensities. The results show the ability to seamlessly combine various styles, indicating potential for use in fields like interior design and producing artistic NFTs.

## VI. FUTURE SCOPE

Future enhancements for the system may include improvements in real-time processing for faster and more efficient style transfer. Through the utilization of effective computational techniques, we can greatly diminish processing time, rendering the system more appropriate for real-life scenarios. Moreover, delving into adaptive style blending using image content and testing out more intricate neural network structures could increase the versatility and excellence of the results. Potential ways to enhance the reach and practicality of this project include adding support for video style transfer and creating a user-friendly interface for manipulating styles.